\documentclass[conference]{IEEEtran}
\IEEEoverridecommandlockouts
\usepackage{cite}
\usepackage{amsmath,amssymb,amsfonts}
\usepackage{algorithmic}
\usepackage{algorithm}
\usepackage{graphicx}
\usepackage{stmaryrd}
\usepackage{xcolor}
\usepackage{array}
\usepackage{commath}
\usepackage{sidecap}
\usepackage{stfloats}
\usepackage{tabularx, boldline}
\usepackage{rotating,booktabs,multirow}
\usepackage{mathtools}
\usepackage{flexisym}
\usepackage{breqn}
\usepackage{url}
\def\R{\mathbb{R}} 
\def\E{\mathbb{E}} 
\newcommand{\mkv}{-\!\!\!\!\minuso\!\!\!\!-}

\usepackage{authblk}
\def\BibTeX{{\rm B\kern-.05em{\sc i\kern-.025em b}\kern-.08em
    T\kern-.1667em\lower.7ex\hbox{E}\kern-.125emX}}

\makeatletter
\newcommand\fs@betterruled{%
  \def\@fs@cfont{\bfseries}\let\@fs@capt\floatc@ruled
  \def\@fs@pre{\vspace*{0.1in}\hrule height.8pt depth0pt \kern2pt}%
  \def\@fs@post{\kern2pt\hrule\relax}%
  \def\@fs@mid{\kern2pt\hrule\kern2pt}%
  \let\@fs@iftopcapt\iftrue}
\floatstyle{betterruled}
\restylefloat{algorithm}
\makeatother

\begin{document}

\title{Deep Directed Information-Based Learning for Privacy-Preserving Smart Meter Data Release}

\author[1]{Mohammadhadi~Shateri}
\author[1]{Francisco~Messina}
\author[2,3]{Pablo~Piantanida}
\author[1]{Fabrice~Labeau}
\affil[1]{\small Department of Electrical and Computer Engineering, McGill University, QC, Canada, \protect\\ Email:\{mohammadhadi.shateri,francisco.messina\}@mail.mcgill.ca}
\affil[2]{Laboratoire des Signaux et Syst\`emes, CentraleSup\'elec-CNRS-Universit\'e Paris Sud, Gif-sur-Yvette, France}
  \affil[3]{Montreal Institute for Learning Algorithms (Mila), Universit\'e de Montr\'eal, QC, Canada}

\maketitle
\thispagestyle{plain}
\pagestyle{plain}
\begin{abstract}
The explosion of data collection has raised serious privacy concerns in users due to the possibility that sharing data may also reveal sensitive information. The main goal of a privacy-preserving mechanism is to prevent a malicious third party from inferring sensitive information while keeping the shared data useful. In this paper, we study this problem in the context of time series data and smart meters (SMs) power consumption measurements in particular. Although Mutual Information (MI) between private and released variables has been used as a common information-theoretic privacy measure, it fails to capture the causal time dependencies present in the power consumption time series data. To overcome this limitation, we introduce the Directed Information (DI) as a more meaningful measure of privacy in the considered setting and propose a novel loss function. The optimization is then performed using an adversarial framework where two Recurrent Neural Networks (RNNs), referred to as the releaser and the adversary, are trained with opposite goals. Our empirical studies on real-world data sets from SMs measurements in the worst-case scenario where an attacker has access to all the training data set used by the releaser, validate the proposed method and show the existing trade-offs between privacy and utility. 
\end{abstract}


\section{Introduction}
In recent decades, there has been an explosive progress in data collection tools that can measure, analyze and disseminate users data. These advances have led to a large impact in several different fields such as electrical power systems, health care, digital banking, etc. However, users are generally unwilling to share their data due to the possibility that a third party could infer their personal private information, e.g., living habits, economic status, health history. Therefore, guaranteeing users privacy while preserving the benefits of data collection is an important challenge in modern data science \cite{nelson2016}. In this paper, we focus on this problem when the data has a time series structure and, in particular, we consider different privacy scenarios motivated by the deployment of Smart Meters (SMs) in electrical distribution networks \cite{giaconi2018privacy}. The SMs are devices that can register a fine-grained electricity consumption of users and communicate this information to the utility provider in almost real-time. The utility of the SMs data is diverse \cite{wang2018review,depuru2011}. It can be used for power quality monitoring, timely fault detection, demand response, energy theft prevention, etc. However, widespread usage of SMs data can lead to serious leakages of consumers' private information, e.g., a malicious third party could use the data to detect the presence of residents at home as well as their personal habits \cite{giaconi2018privacy}. This problem can have a serious impact on the deployment pace of SMs and, more broadly, in the development of smart electrical grids. Thus, it is critical to ensure that SMs data are sanitized before being released.

A very simple strategy that has been proposed in the context of SMs is to use pseudonyms rather than the real identities of users for data publishing purposes \cite{efthymiou2010smart}. However, this approach implicitly assumes that a trusted anonymizer is available. Another simple technique suggested in the literature is downsampling of the data, where the sampling rate is reduced to a level that does not pose any privacy threat \cite{cardenas2012privacy,mashima2015authenticated}. Although this approach may be effective from the privacy point of view, it could also limit seriously  the utility of the SMs data for some applications requiring a timely response. More sophisticated and recent approaches exploit the presence of renewable energy sources and rechargeable batteries in homes to modify the actual energy consumption of users in order to hide the sensitive information \cite{backes2013differentially,zhao2014achieving,li2018information,giaconi2018privacy,erdemir2019privacy}. Some of these works use ideas from the well-known principle of differential privacy. However, recent articles suggest that the utility loss of differential privacy may be significant in practice \cite{mendes2017privacy}. It should be noted that our approach to the problem, which works only with the power measurement data, does not preclude the use of methods that change the energy consumption patterns by using physical resources. In fact, it should be viewed as a complementary approach that could even be used on top of the above mentioned methods.

In an information-theoretic context, privacy is generally measured by the Mutual Information (MI) between the sensitive and release variables \cite{li2018information,giaconi2018privacy,erdemir2019privacy,sankar2013smart}. Some of these studies aim to find a privacy-utility trade-off using ideas from rate-distortion theory \cite{sankar2013smart,tripathy2017privacy}. More specifically, the theoretical framework of the privacy-utility problem was proposed in \cite{sankar2013smart}, where a hidden Markov model for the power measurements of SMs is considered in which the distribution is assumed to be controlled by only the state of the home appliances. The privacy-utility trade-off is then found for a stationary Gaussian model of electricity load with MI between release and private sequence of variables as a privacy measure. Besides the limitation of the Gaussian model, it is noted that the MI is not well-suited to capture the causal structure of the time series data.

In this paper, an information-theoretic cost function for privacy-preserving data release of time series is proposed. In order to take into account the time series structure and causality of the data in the privacy measure, we use the Directed Information (DI) \cite{massey1990causality} between the sensitive time series and an estimation of it. Then, a cost function is derived for the releaser mechanism based on an upper bound of the DI. To optimize and validate our cost function without imposing constraints on the data distribution, two recurrent neural networks, named as releaser and adversary networks, are employed. This approach is based on the framework of Generative Adversarial Networks (GANs) \cite{goodfellow2014,huang2017context}, where two neural networks are trained simultaneously with opposite goals. We will show that by controlling the relative weight between a distortion measure and the DI privacy measure, we can control the utility-privacy trade-off of SMs power measurements. A similar approach for the privacy problem, but for different applications, was considered in \cite{tripathy2017privacy,2018arXiv180209386F} considering the standard MI and independent and identically distributed (i.i.d.) data, where the authors use two deep feed-forward neural networks for the releaser and adversary. However, to the best of our knowledge, this is the first work to consider a DI privacy measure for time series data in the general privacy-preserving context and in SMs applications in particular.

This paper is organized as follows. In Section \ref{sec:formulation}, we present the theoretical formulation of the problem. Then, in Section \ref{sec:model}, a privacy-preserving data release method based on Long-Short Term Memory (LSTM) Recurrent Neural Networks (RNNs) is introduced along with the training algorithm. Results for two different applications based on SMs data are presented in Section \ref{sec:results}. Finally, some concluding remarks and a discussion about future work are given in Section \ref{sec:conclusion}.

\subsection*{Notation and conventions}
A sequence of random variables $(X_1, \ldots, X_T)$ of length $T$ is denoted by $X^T$, while $x^T = (x_1, x_2, \ldots, x_T)$ denotes a realization of $X^T$ and $x^{(i)T} = (x^{(i)}_1, x^{(i)}_2, \ldots, x^{(i)}_T)$ denotes the $i^{\text{th}}$ sample in a minibatch used for training. Mutual information~\cite{cover2006elements} between variables $X$ and $Y$ is represented as $I(X;Y)$ and the entropy as $H(X)$. We use ${X \mkv Y \mkv Z}$ to indicate that $X$, $Y$ and $Z$ form a Markov chain. The expectation of a random variable $X$ is denoted as $\E[X]$. 

\section{Problem Formulation and Training Objective}
\label{sec:formulation}

\subsection{Main definitions}

Consider the private variables $X^T$ (such as occupancy label, household identity, or acorn family type), useful variables $Y^T$ (such as actual electricity consumption of household), and observed variables $W^T$ (which could be a combination of private and useful variables). We assume that $X_t$ takes values on a discrete alphabet $\mathcal{X}$, for $t \in \{ 1, \ldots, T \}$. A releaser $\mathcal{R}_{\theta}$ (this notation is used to denote that the releaser is controlled by its parameters $\theta$) produces the release variables as $Z_t$ based on the observation $W^t$, for each time $t \in \{ 1, \ldots, T \}$, while an adversary $\mathcal{A}_{\phi}$ attempts to infer $X_t$ based on $Z^t$ by finding an approximation of $p_{X^T|Z^T}$ which we shall denote by $p_{\hat{X}^T|Z^T}$. Thus, the Markov chain $(X^t,Y^t) \mkv W^t \mkv Z^t \mkv \hat{X}^t$ holds for all $t \in \{ 1, \ldots, T \}$. In addition, due to causality, the distribution $p_{Z^T\hat{X}^T|W^T}$ can be decomposed as follows:
\begin{equation} p_{Z^T\hat{X}^T|W^T}(z^T,\hat{x}^T|w^T) = \prod_{t=1}^{T} p_{Z_t|W^t}(z_t|w^t) p_{\hat{X}_t|Z^t}(\hat{x}_t|z^t). \end{equation}

The goal of the releaser $\mathcal{R}_\theta$ is to minimize the flow of information from the sensitive variables $X^T$ to their estimation $\hat{X}^T$ while simultaneously keeping the distortion between the release variables $Z^T$ and the useful variables $Y^T$ below some given value. On the other hand, the goal of the adversary $\mathcal{A}_{\phi}$ (this notation is used to denote that the adversary is controlled by its parameters $\phi$) is to estimate $X^T$ as accurately as possible. 

To take into account the causal relation between $X^T$ and $\hat{X}^T$, the flow of information is quantified by the DI~\cite{massey1990causality}: 
\begin{equation} \label{eqtr1} I\big(X^T\rightarrow \hat{X}^T\big)=\sum_{t=1}^{T} I(X^t;\hat{X}_t|\hat{X}^{t-1}), \end{equation}
where $I(X^t;\hat{X}_t|\hat{X}^{t-1})$ is the conditional mutual information between $X^t$ and $\hat{X}_t$ conditioned on $\hat{X}^{t-1}$\cite{cover2006elements}.

The expected distortion between $Z^T$ and $Y^T$ is defined as:
\begin{equation} \mathcal{D}(Z^T,Y^T) \triangleq \E[d(Z^T,Y^T)], \end{equation}
where $d : \R^T \times \R^T \to \R$ is any distortion function (i.e., a metric on $\R^T$). In order to ensure the quality of the release we shall impose the following constraint: $\mathcal{D}(Z^T,Y^T) \le \varepsilon$ for some given $\varepsilon \ge 0$. In this work, we will consider the normalized squared error as in \cite{sankar2013smart}, i.e., \begin{equation} d(z^T, y^T) \triangleq \frac{1}{T} \sum_{t=1}^T (z_t-y_t)^2. \end{equation}
Nevertheless, it should be noted that other distortion measures can also be relevant for the SMs data. For instance, demand response programs usually require an accurate knowledge of peak power consumption, so a distortion function closer to the infinity norm would be more meaningful for this particular application. This brief discussion simply illustrates that the distortion function should be properly matched to the intended application of the release variables $Z^T$ in order to preserve the characteristics of the useful variables $Y^T$ that are considered essential. Since the goal of this paper is mainly to introduce a new privacy measure and privacy-preserving data release framework, we will not further investigate different fidelity measures.

Therefore, the problem of finding an optimal releaser subject to the aforementioned adversary and distortion constraint can be formally written as follows:
\begin{align} \label{eqtr2} & \underset{\theta}{\text{min}} \; \quad I\left(X^T\rightarrow \hat{X}^T\right), \nonumber \\ & \text{s.t. } \quad \mathcal{D}(Z^T,Y^T) \le \varepsilon. \end{align}
Note that the solution of this optimization problem is a function of $p_{\hat{X}^T|Z^T}$, the conditional distributions that represent the adversary $\mathcal{A}_{\phi}$.

\subsection{Novel training objective}

The optimization problem \eqref{eqtr2} can be directly used to define a loss function for $\mathcal{R}_\theta$. However, note that the cost of computing the DI term is $O(|\mathcal{X}|^T)$, where $|\mathcal{X}|$ is the size of $\mathcal{X}$. Thus, for the sake of tractability, DI will be replaced with the following surrogate bound:
\begin{align} \label{eqtr5}
I\left(X^T\rightarrow \hat{X}^T\right) & = \sum_{t=1}^{T}H(\hat{X}_t|\hat{X}^{t-1})-H(\hat{X}_t|\hat{X}^{t-1},X^t)
\nonumber \\  &\overset{\text{(i)}}{\leq}\sum_{t=1}^{T}H(\hat{X}_t|\hat{X}^{t-1})-H(\hat{X}_t|\hat{X}^{t-1},X^t,Z^t) \nonumber \\ 
&\overset{\text{(ii)}}{=} \sum_{t=1}^{T}  H(\hat{X}_t|\hat{X}^{t-1})-H(\hat{X}_t|Z^t)\nonumber \\ 
& \overset{\text{(iii)}}{\leq} T \log|\mathcal{X}| - \sum_{t=1}^{T}H(\hat{X}_t|Z^t), 
\end{align}
where (i) is due to the fact that conditioning reduces entropy; equality (ii) is due to the Markov chains $X^t \mkv Z^t \mkv \hat{X}^t$ and $\widehat{X}^{t-1} \mkv Z^t \mkv \widehat{X}^t$; and (iii) is due to the trivial bound $H(\hat{X}_t|\hat{X}^{t-1}) \le H(\hat{X}_t) \le \log |\mathcal{X}|$. Therefore, the loss function for $\mathcal{R}_\theta$ can be written as
\begin{equation} \label{eq:releaser_loss} 
\mathcal{L}_{\mathcal{R}}(\theta, \phi,\lambda) =  \mathcal{D}(Z^T,Y^T) - \frac{\lambda}{T} \sum_{t=1}^{T} H(\hat{X}_t|Z^t) , 
\end{equation}
where $\lambda \ge 0$ controls the privacy-utility trade-off and the factor $1/T$ has been introduced for normalization purposes. 
It should be noted that the value of $\lambda$ in \eqref{eq:releaser_loss} indirectly controls the achievable $\varepsilon$ in \eqref{eqtr2}, which means that we can control the privacy-utility trade-off by varying $\lambda$.
For $\lambda = 0$, the loss function $\mathcal{L}_{\mathcal{R}}(\theta,\phi,\lambda)$ reduces to the expected distortion, being independent from the adversary $\mathcal{A}_{\phi}$. In such scenario, $\mathcal{R}_\theta$ offers no privacy guarantees. Conversely, for very large values of $\lambda$, the loss function $\mathcal{L}_{\mathcal{R}}(\theta,\phi,\lambda)$ is dominated by the upper bound on the DI, so that privacy is the only goal of $\mathcal{R}_\theta$. In this regime, we expect the adversary $\mathcal{A}_{\phi}$ to completely fail in the task of estimating $X^T$, i.e., to approach to random guessing performance.

On the other hand, the adversary $\mathcal{A}_{\phi}$ is a classifier which optimizes the following cross-entropy loss:
\begin{equation} \label{eq:attacker_loss}
\mathcal{L}_{\mathcal{A}}(\phi) = \frac{1}{T} \sum_{t=1}^T \E\left[- \log p_{\hat{X}_t|Z^t}(X_t|Z^t) \right], 
\end{equation}
where the expectation should be understood w.r.t. $p_{X_tZ^t}$. Notice that
\begin{equation} \frac{1}{T} \sum_{t=1}^{T} H(X_t|Z^t)\leq\mathcal{L}_{\mathcal{A}}(\phi). \end{equation}
Therefore, if the adversary is ideal (i.e., $p_{\hat{X}_t|Z^t} = p_{X_t|Z^t}$ for all $t$), the releaser network, by maximizing $\frac{1}{T} \sum_{t=1}^{T} H(\hat{X}_t|Z^t)$, prevents the adversary to infer private data.


\section{Privacy-Preserving Mechanism} \label{sec:model}

\begin{figure}[t!]
	\centering
	\includegraphics[width=1\linewidth]{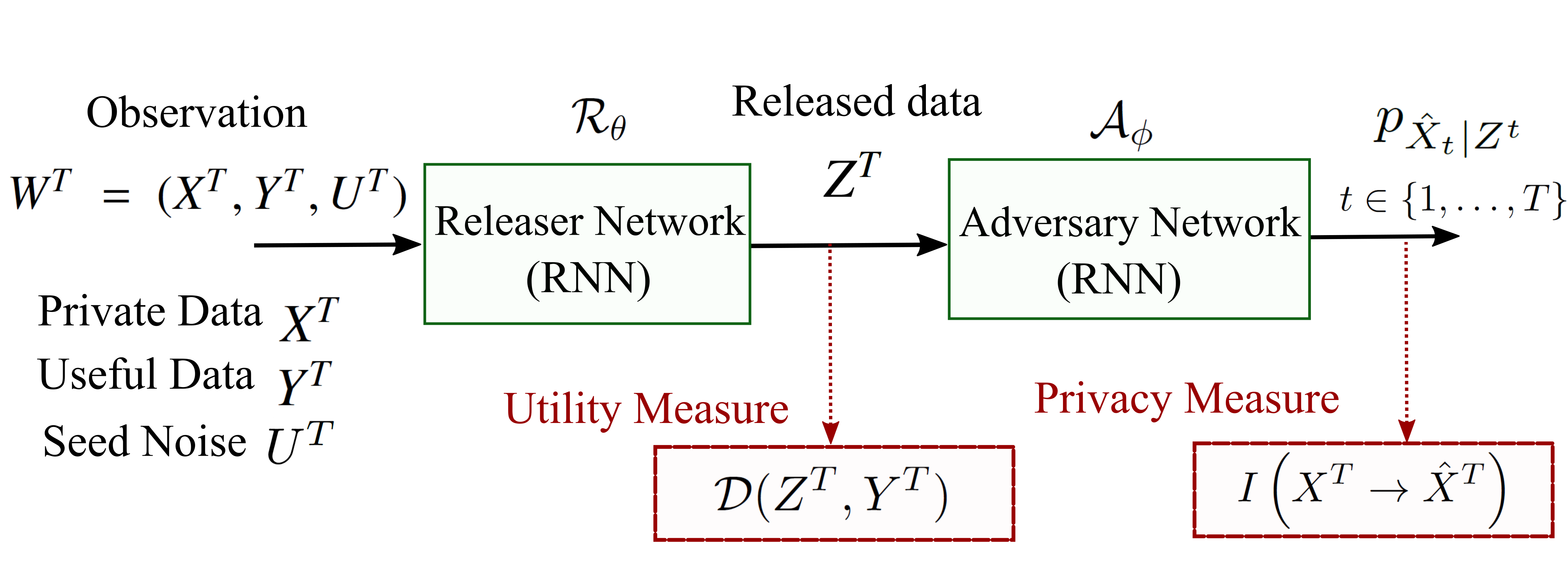}
	\caption{Privacy-Preserving framework. The seed noise $U^T$ is generated from i.i.d. samples according to a uniform distribution: $U_t \sim U[0,1]$.}
	\label{had1}
\end{figure}

\begin{figure}[htbp]
	\centering
	\includegraphics[width=1\linewidth]{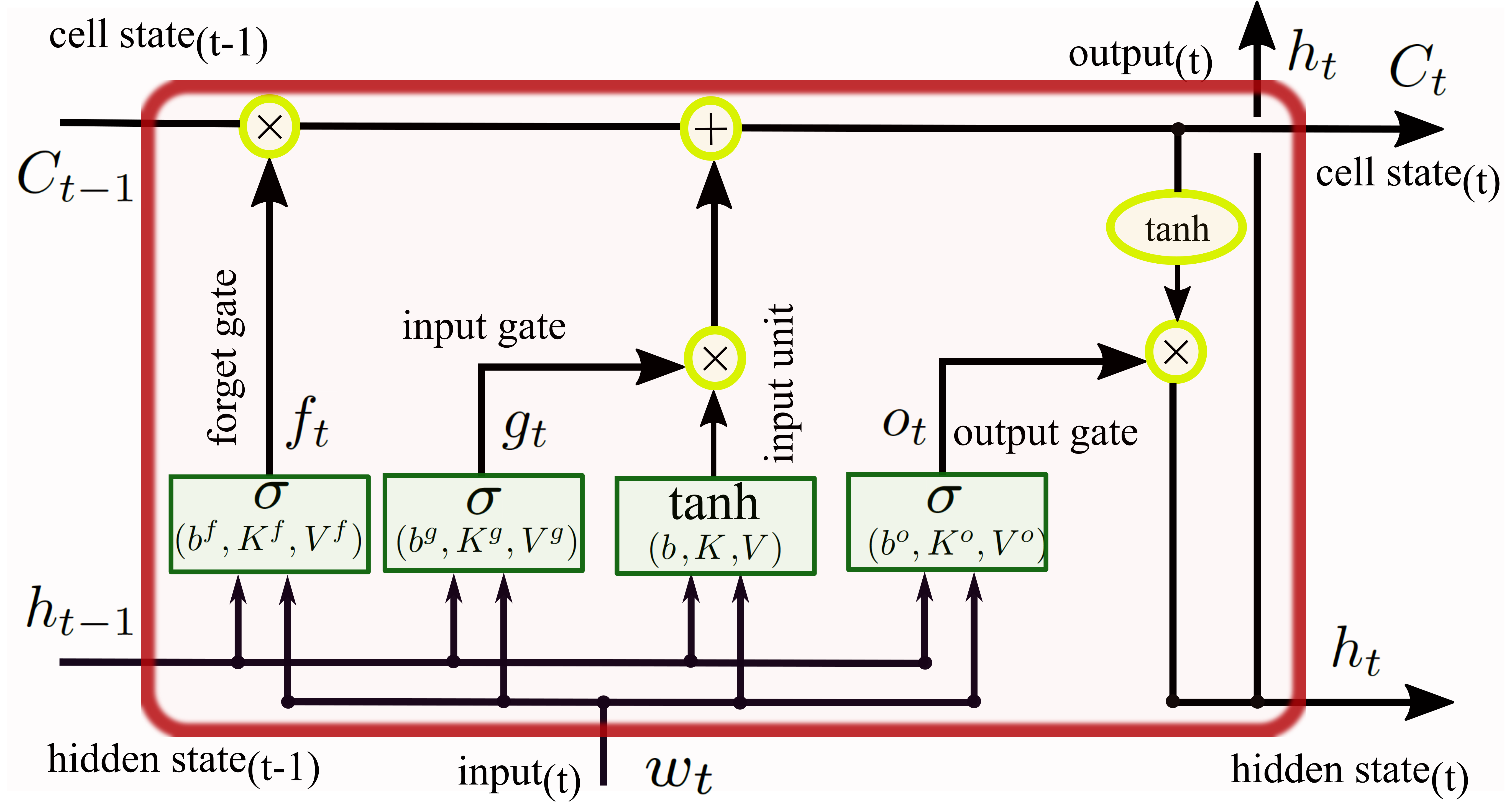}
	\caption{LSTM recurrent network cell diagram. The cell includes four gating units to control the flow of information. All the gating units have a sigmoid activation function ($\sigma$) except for the input unit (that uses an hyperbolic tangent activation function ($\tanh$) by default). The parameters $b,V,W$ are respectively biases, input weights, and recurrent weights. In the LSTM architecture, the forget gate $f_t = \sigma(b^f + K^fh_{t-1} + V^fw_t)$ uses the output of the previous cell (which is called hidden state $h_{t-1}$) to control the cell state $C_t$ to remove irrelevant information. On the other hand, the input gate $g_t = \sigma(b^g + K^g h_{t-1} + V^g w_t)$ and input unit adds new information to $C_t$ from the current input. Finally, the output gate $o_t = \sigma(b^o + K^oh_{t-1} + V^ow_t)$ generates the output of the cell from the current input and cell state.}
	\label{had2}
\end{figure}

Based on the previous theoretical formulation, an adversarial modeling framework consisting of two RNNs, a releaser $\mathcal{R}_{\theta}$  and an adversary $\mathcal{A}_{\phi}$, is considered (see Fig. \ref{had1}). Note that independent noise $U^T$ is appended to $W^T$ in order to randomize the released variables $Z^T$, which is a popular approach in privacy-preserving methods. In addition, the available theoretical results show that, for Gaussian distributions, the optimal release contains such a noise component \cite{sankar2013smart,tripathy2017privacy}. For both networks, a LSTM architecture is selected (see Fig. \ref{had2}), which was shown to be successful in several problems dealing with sequences of data (see \cite{goodfellow2016} and references therein for more details). The training of the suggested framework is performed using Algorithm \ref{Al1} which uses $k$ gradient steps to train $\mathcal{A}_\phi$ followed by one gradient step to train $\mathcal{R}_\theta$. Note that $k$ should be large enough to ensure that $\mathcal{A}_\phi$ is a strong adversary during training. This is in fact the common practice for effectively training two networks in an adversarial framework \cite{goodfellow2014} and, in particular, in privacy scenarios \cite{tripathy2017privacy}. It should be recalled that, after the training of both networks is completed, an attacker network is trained in order to test the privacy achieved by the releaser network. It should be clarified that this attacker network is distinct from the adversary network used during training and illustrated in Fig~\ref{had1}: the attacker used in testing mimics a real-world attacker that would try to deduce the private date from the release data. 

\begin{algorithm*}
	\caption{Algorithm for training privacy-preserving data releaser neural network.}
	\label{Al1}
	\textbf{Input:} Data set (which includes sample sequences of useful data $y^T$, sensitive data $x^t$); seed noise samples $u^T$; seed noise dimension $m$; batch size $B$; number of steps to apply to the adversary $k$; gradient clipping value $C$; $L_2$ recurrent regularization parameter $\beta$. \\
	\textbf{Output:} Releaser network $\mathcal{R}_\theta$.
	\begin{algorithmic}[1]
		\FOR {number of training iterations}
	
		\FOR {$k$ steps}
		\STATE Sample minibatch of $B$ examples: $\mathcal{B} = \Big\{w^{(b)T}=\Big(x^{(b)T},y^{(b)T},u^{(b)T}\Big); \; b=1,2,..,B\Big\}$.
		\STATE Compute the gradient of $\mathcal{L}_{\mathcal{A}}(\phi)$, approximated with the minibatch $\mathcal{B}$, w.r.t. to $\phi$.
		\STATE Update the adversary by applying the RMSprop optimizer with clipping value $C$.
		\ENDFOR
		
		\STATE Sample minibatch of $B$ examples: $\mathcal{B} = \Big\{w^{(b)T}=\Big(x^{(b)T},y^{(b)T},u^{(b)T}\Big); \; b=1,2,..,B\Big\}$.	\STATE Compute the gradient of $\mathcal{L}_{\mathcal{R}}(\theta,\phi,\lambda)$, approximated with the minibatch $\mathcal{B}$, w.r.t. to $\theta$. 
		\STATE Use $\textrm{Ridge}(L_2)$ recurrent regularization with value $\beta$ and update the releaser by applying RMSprop optimizer with clipping value $C$.
		\ENDFOR
	\end{algorithmic}
\end{algorithm*}

\section{Results and Discussion} \label{sec:results}


\subsection{Description of datasets}

In this study, the Electricity Consumption \& Occupancy (ECO) and Pecan Street data sets are used. The ECO data set, collected and published by \cite{beckel2014eco}, includes 1 Hz power consumption measurements and occupancy information of five houses in Swiss over a period of $8$ months. In this study we re-sampled the data to have hourly samples. On the other hand, the Pecan Street data set contains hourly SMs data of houses in Texas, Austin and was collected by Pecan Street Inc. \cite{street2019dataport}. Pecan Street project is a smart grid demonstration research program which provides electricity, water, natural gas, and solar energy generation measurements for over $1000$ houses in Texas, Austin. In order to model time dependency over each day (with a data rate of 1 sample per hour), the data was reshaped to sample sequences of length $24$. For the ECO and Pecan Street data set, a total number of $11225$ and $9120$ sample sequences are used, respectively. The data is splitted into train and test sets with a ratio of roughly 85 : 15 while $10 \%$ of the training data is used as the validation set. It should be noted that in this study, we assume that the attacker has access to all the training data used by the releaser, which can be considered as a worst-case scenario study.

\subsection{Inference of households occupancy} \label{sec:household_occupancy}

The first practical case of study regarding privacy-preserving in time series data is the concern of inferring presence/absence of residents at home from the total power consumption collected by SMs \cite{kleiminger2015household,jia2014human}. For this application, the electricity consumption measurements from the ECO data set are considered as the useful data, while occupancy labels are considered as the private data. Therefore, our privacy-preserving data release method aims to minimize a trade-off between the distortion of the total electricity consumption incurred and the probability of inferring the presence of an individual at home from the release signal. The releaser and adversary networks used for the training consist of 4 LSTM layers with $64$ cells and 2 LSTM layers with $32$ cells, respectively where a $\tanh$ activation function used. In addition, recurrent regularizer with parameter $\beta = 1.5$ was used in each layer of the release network. The values of the other hyperparameters ($B$, $k$,$m$) were set to $(128,4,8)$, respectively. Finally, after training, a strong attacker is used, consisting of 3 LSTM layers. 
\begin{figure}[htbp]
    \vspace{0.04in}
	\centering
	\includegraphics[width=.9\linewidth]{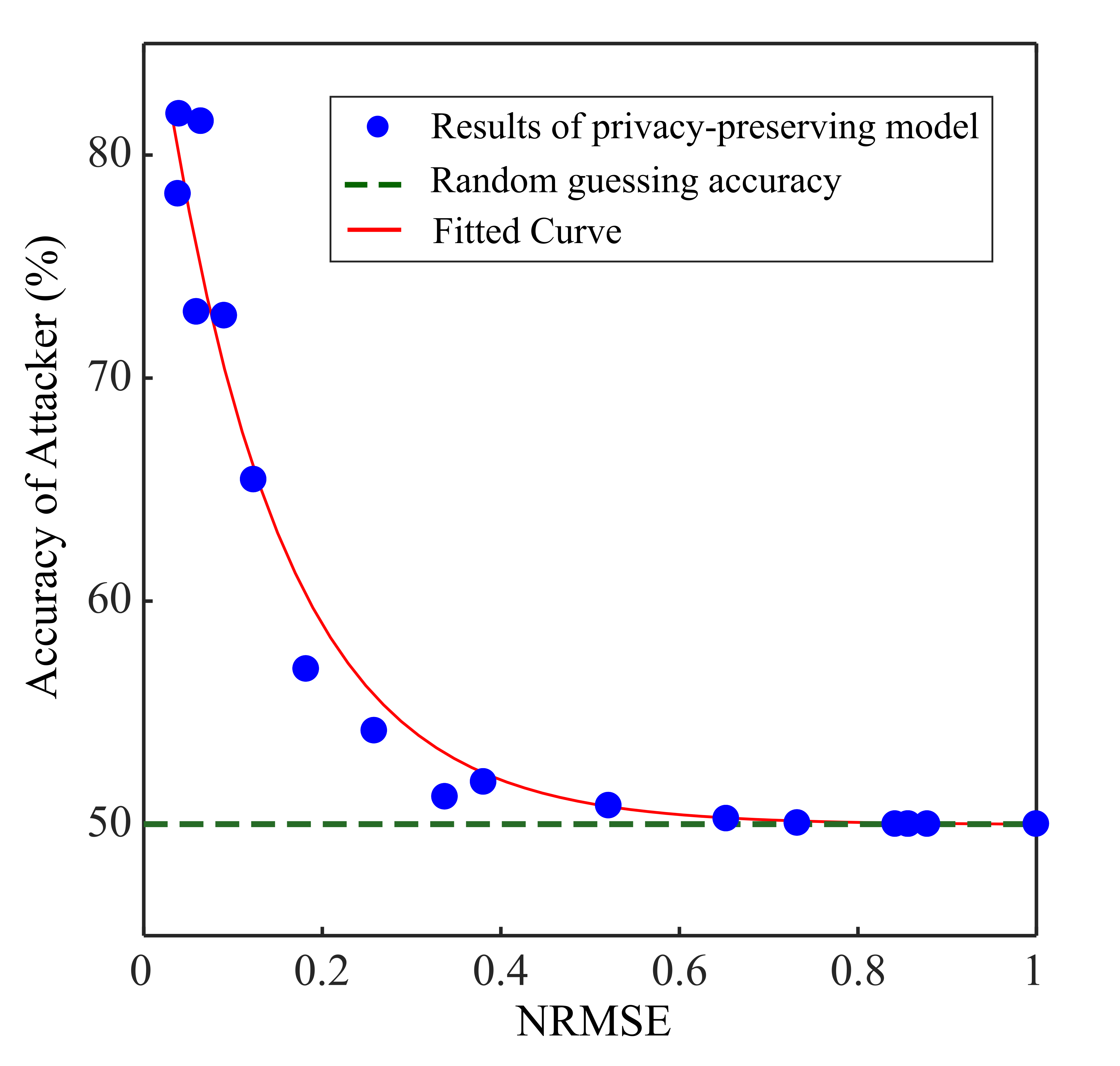}
	\caption{Privacy-utility trade-off for house occupancy inference application. Since in this application the attacker is a binary classifier, the random guessing (balanced) accuracy is 50$\%$. The fitted curve is based on an exponential function and is included only for illustration purposes.}
	\label{had3}
\end{figure}


Based on the target data $Y^T$ and the release data $Z^T$, the normalized root mean-square-error (NRMSE) is defined by
\begin{equation} \label{eq:NormalizedRMSE} \text{NRMSE} \triangleq \sqrt{ \frac{ \E\left[\norm{Y^T-Z^T}^2\right]}{\E\left[\norm{Y^T}^2\right]}}. \end{equation}
Fig. \ref{had3} shows the empirically found privacy-utility trade-off for this application. It can be seen that by adding more distortion on the released data, the attacker is pushed toward a random guessing classifier.

\begin{figure}[htbp]
    \vspace{0.04in}
	\centering
	\includegraphics[width=0.9\linewidth]{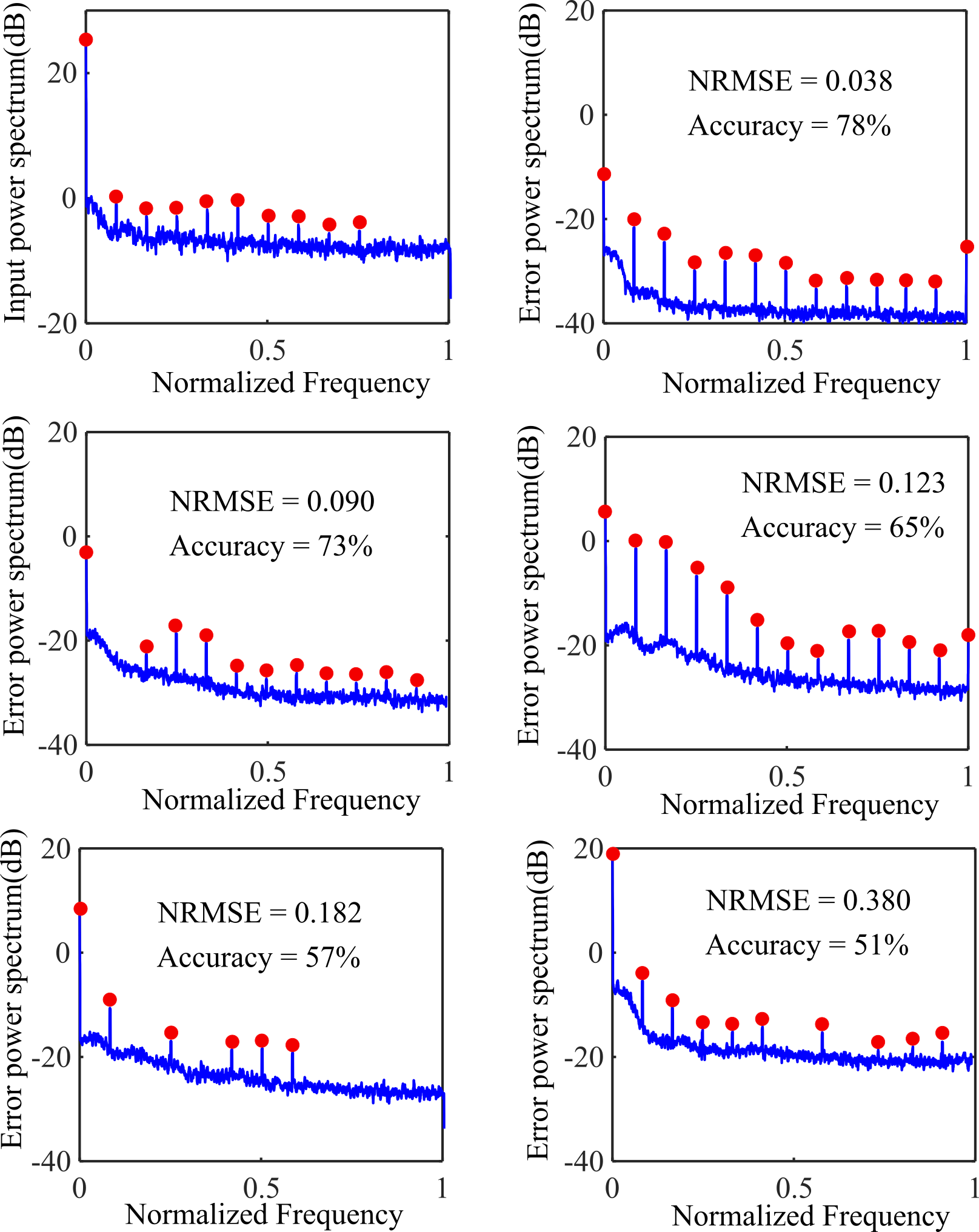}
	\caption{PSD of the actual electricity consumption and error signals for the house occupancy inference application.}
	\label{had4}
\end{figure}

In order to provide more insights about the release mechanism, the Power Spectrum Density (PSD) of the input signal and the PSD of the error signal (defined as the difference between the actual power consumption and the released signal) for four different cases along the privacy-utility trade-off curve of Fig.~\ref{had3} are estimated using Welch's method \cite{stoica2005spectral}. For each case, we use 10 release signals and average the PSD estimates. Results are shown in Fig. \ref{had4}. Looking at the PSD of the input signal (useful data) some harmonics are visible. The PSD of the error signals show that the model controls the trade-off in privacy-utility by mainly modifying the distortion on these harmonics. 

It should be mentioned that two stationary tests including the Augmented Dickey-Fuller test \cite{dickey1979distribution} and the Kwiatkowski, Phillips, Schmidt, and Shin (KPSS) test\cite{kwiatkowski1992testing} applied to our data set indicates that there is enough evidence to suggest the data is stationary, supporting our PSD analysis. 

\subsection{Inference of house identity}

\begin{figure}[htbp]
	\centering
	\includegraphics[width=0.8\linewidth]{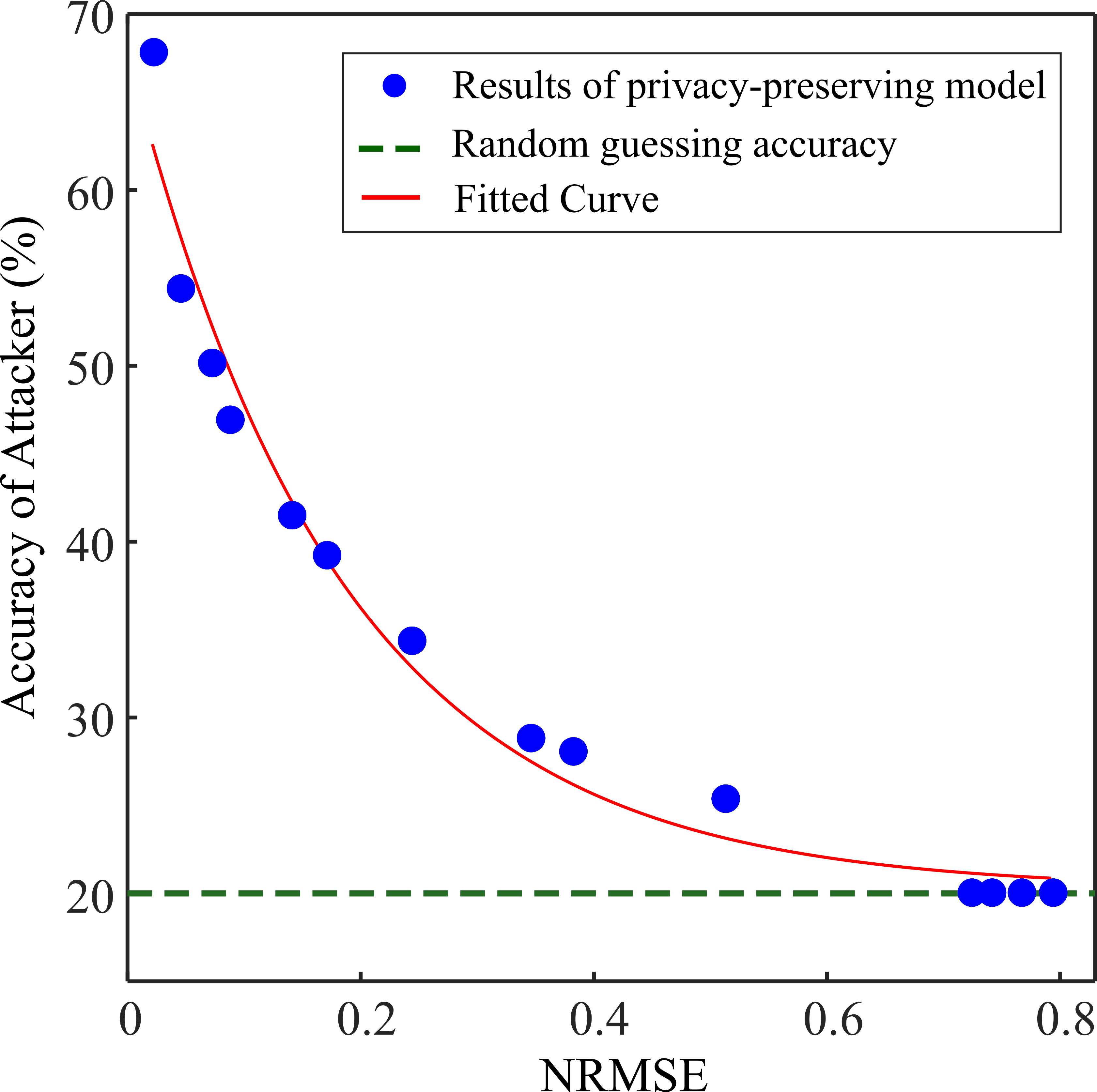}
	\caption{Privacy-utility trade-off for house identity inference application. Since in this application the attacker is a five-class classifier, the random guessing (balanced) accuracy is 20$\%$. The fitted curve is based on an exponential function and is included only for illustration purposes.}
	\label{had5}
\end{figure}

The second practical case of study regarding the privacy-preserving in SMs measurements is identity recognition from total power consumption of households \cite{efthymiou2010smart}. It is assumed that the attacker has access to total power consumption of different households in a region (training data) and then attempts to determine identities of the households using the new released data (test data). Thus, our model aims at generating release data of total power consumption of households in a way that prevents the adversary to perform the identity recognition while keeping distortion on the total power minimized. For this task, total power consumption of five houses is used. For this application, the releaser consists of 6 LSTM layers each includes $128$ cells and adversary has 4 LSTM layers with $32$ cells. Similarly, a $\tanh$ activation function was applied and $\beta = 2$ was used in each layer of the release network. The values of the other hyperparameters ($B$, $k$,$m$) are set to $(128,5,3)$, respectively. Finally, after training, an attacker, consisting of 4 LSTM layers was used. The empirical privacy-utility trade-off curve obtained for this application is presented in Fig. \ref{had5}. Comparing Fig. \ref{had5} with Fig. \ref{had3}, we see that a high level of privacy is expensive. For instance, in order to obtain an attacker accuracy of 30 $\%$, the NRMSE should be approximately equal to 0.30. This is attributed to the fact that this task is harder from the learning point of view than the one considered in Section \ref{sec:household_occupancy}.


\begin{figure}[htbp]
	\centering
	\includegraphics[width=0.9\linewidth]{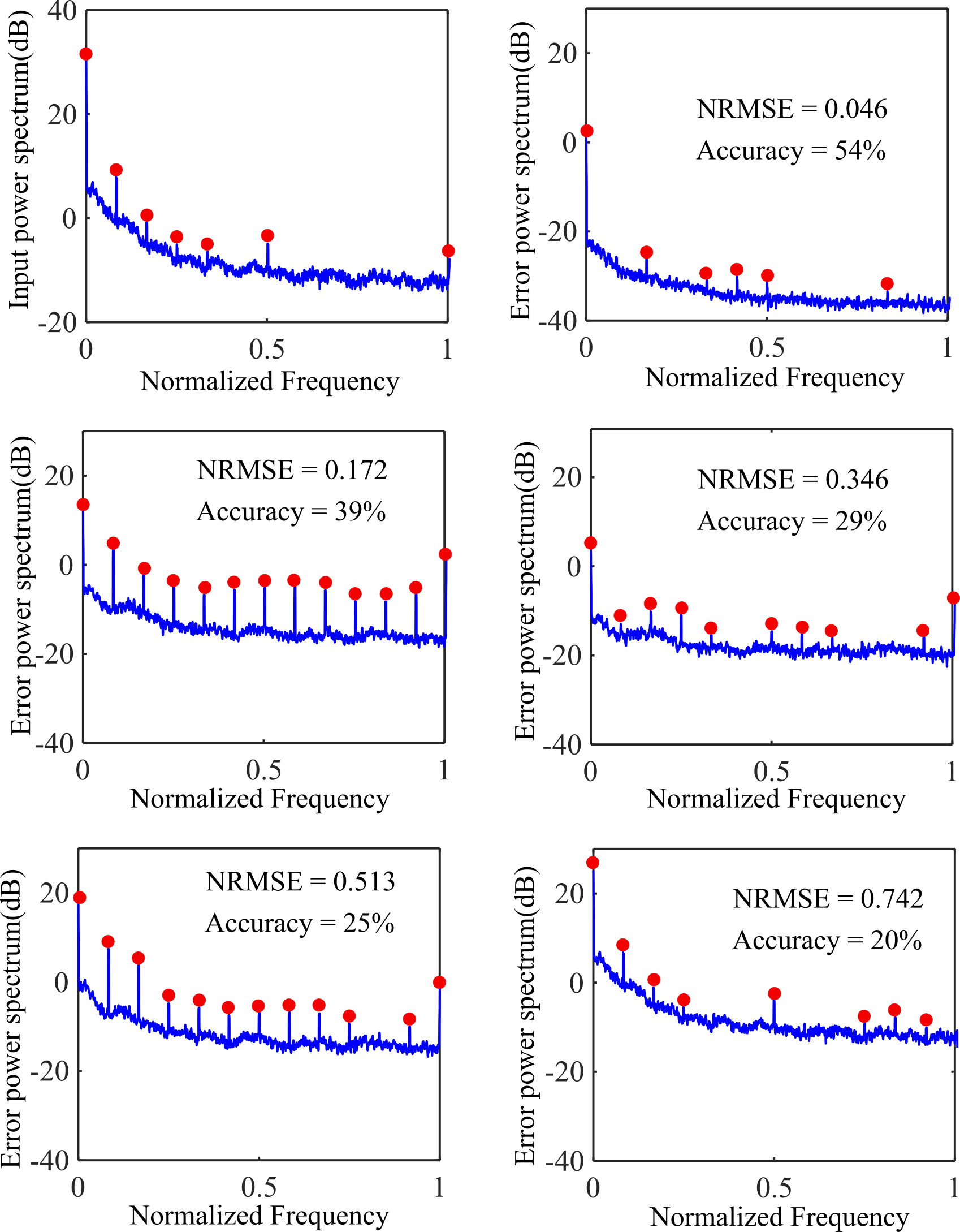}
	\caption{PSD of the actual electricity consumption and error signals for the house identity inference application.}
	\label{had6}
\end{figure}

PSD analysis was also performed for this application, yielding the results of Fig. \ref{had6}. Once again, we see that the release network provides privacy-utility trade-off by mainly distorting the harmonics on the actual electricity consumption signal.

\section{Conclusion} \label{sec:conclusion}

We have presented a new method to train privacy-preserving mechanisms controlling the privacy-utility trade-off in time series data. This lead us to define the directed information between sensitive variables and their estimation as a more suitable privacy measure than previous proposals in the literature. A tractable upper bound was then derived and a deep learning adversarial framework between two recurrent neural networks was introduced to optimize the new loss function. Our method was validated with two well-known privacy problems in smart meters data using two different open data sets. For both privacy problems we considered the worst-case where an attacker has access to all the training data used by the releaser. In future work, we will consider alternative formulations of the problem such as different distortion measures and a more general loss function in order to attempt to provide universal privacy guarantees (i.e., independent of the attacker structure and computational power).


\section*{Acknowledgment}

This work was supported by Hydro-Quebec, the Natural Sciences and Engineering Research Council of Canada, and McGill University in the framework of the NSERC/Hydro-Quebec Industrial Research Chair in Interactive Information Infrastructure for the Power Grid (IRCPJ406021-14). This project has received funding from the European Union’s Horizon 2020 research and innovation programme under the Marie Skłodowska-Curie grant agreement No 792464.

\bibliographystyle{ieeetr}
\bibliography{HREF}

\end{document}